\documentclass{article}

\usepackage[final]{nips_2018}

\usepackage[utf8]{inputenc} % allow utf-8 input
\usepackage[T1]{fontenc}    % use 8-bit T1 fonts
\usepackage{hyperref}       % hyperlinks
\usepackage{url}            % simple URL typesetting
\usepackage{booktabs}       % professional-quality tables
\usepackage{amsfonts}       % blackboard math symbols
\usepackage{nicefrac}       % compact symbols for 1/2, etc.
\usepackage{microtype}      % microtypography
\usepackage{graphicx}       % Inclusion of graphics

\title{On the role of neurogenesis in overcoming \\ catastrophic forgetting}

\author{
  German I. Parisi\\
  Apprente, Inc.\\
   Mountain View, CA, USA\\
  \texttt{german@apprente.com} \\
  \And
  Xu Ji \\
  University of Oxford \\
  Oxford, UK \\
  \texttt{xuji@robots.ox.ac.uk} \\
  \And
  Stefan Wermter \\
  University of Hamburg \\
  Hamburg, Germany \\
  \texttt{wermter@informatik.uni-hamburg.de} \\
}

\begin{document}

\maketitle

\begin{abstract}
Lifelong learning capabilities are crucial for artificial autonomous agents operating on real-world data, which is typically non-stationary and temporally correlated.
In this work, we demonstrate that dynamically grown networks outperform static networks in
incremental learning scenarios, even when bounded by the same amount of memory in both cases. Learning is unsupervised in our models, a condition that additionally makes training more challenging whilst increasing the realism of the study, since humans are able to learn without dense manual annotation.
Our results on artificial neural networks reinforce that structural plasticity constitutes
effective prevention against catastrophic forgetting in non-stationary environments, as well as empirically supporting the importance
of neurogenesis in the mammalian brain.
\end{abstract}

\section{Introduction}

Artificial agents and autonomous robots should be able to learn in a lifelong manner, continually acquiring and fine-tuning knowledge through interaction with the environment~[1].
Since real-world data is naturally non-stationary - for example, video
may focus on different scenes or objects in adjacent temporal blocks - it is desirable for learning algorithms to operate on non-stationary input distributions without suffering from \textit{catastrophic forgetting}~[2], i.e. forgetting modes of the function that had previously been retained.
For instance, if a trained object classification model is subsequently shifted to a different set of objects, it would proceed to forget the original set due to a representation shift, which is quantitatively observable via decreasing classification accuracy on previously learned classes.
In addition to catastrophic forgetting, the requirement for dense human annotations used by standard regression techniques is also undesirable as such labels are typically not present in real-world learning scenarios.
Furthermore, since storing previously encountered data samples is expensive, learning algorithms should be able to replay learned schema to consolidate internal representations, allowing for continual learning in the absence of external sensory stimulation~[3].

The mammalian brain remains the best model of unsupervised incremental learning, which makes biologically-inspired learning algorithms a compelling approach.
Proposed solutions for mitigating catastrophic interference in this domain have focused on regulating intrinsic levels of plasticity to protect acquired knowledge given a fixed number of neurons~[5,~6,~13], dynamically allocating new neurons to accommodate novel dissimilar knowledge~[7,~8,~16], and sample replay to simulate a stationary distribution in its absence~[9,~15].
The general notion of structural plasticity is widely used across machine learning literature, for example in fine-tuning learned networks by appending new layers [16], making it a promising solution to lifelong learning in its own right, even when disregarding biological desiderata.
In previous work~[9], we introduced a self-organizing neural architecture which combines the latter two ideas, updating synaptic connectivity patterns using an unsupervised competitive Hebbian learning rule, augmenting input via memory replay, and growing new neurons when activation levels fall short of an empirically defined neural activity threshold.
The growing self-organizing network is composed of a dynamic set of recurrent neurons for learning the spatiotemporal structure of input sequences and has been successfully applied to incremental object and action recognition from videos~[8,~9].
However, the question arises whether the process of neurogenesis functionally contributes to achieving better performance or if the same performance could be achieved by the use of sufficiently large networks in similar experimental set-ups.

In this work, we disentangle the factor of neurogenesis in mitigating catastrophic forgetting.
We test GWR on both static and growing architectures, with the capacity of the latter bounded by the size of the former for fairness, otherwise any improvements in performance could be attributed to a greater memory capacity.
Allowing neurogenesis results in an accuracy improvement on the CORe50 dataset~[10] for incrementally learned object recognition, compared to a static network with memory equal to the maximum bound, showing that neurogenesis functionally contributes to preventing forgetting.
The network trained with neurogenesis was also able to retain newly trained classes faster, which is particularly interesting given that the \textit{stability-plasticity dilemma} suggests lower retention rates are required to mitigate forgetting~[4], as fast retention contravenes representation stability.

\section{Growing Self-Organizing Memory}

Neurogenesis and synaptic rewiring via structural plasticity play a crucial role in the formation of memory from episodic experience~[11].
A biological function of structural plasticity is to increase information storage efficiency in terms of space and energy demands.
The definition of storage capacity is the number of memory associations represented by activity patterns at different times, so that memories are identified with patterns of neural activity that can be updated, stabilized, and evoked by modified synaptic connectivity.
From a modeling perspective, neural networks that grow neurons and develop connectivity patterns over time have the advantage of mitigating the disruptive interference of existing internal representations when learning from novel sensory observations.

Here, we focus on growing recurrent self-organizing networks that learn the spatiotemporal structure of the input in an unsupervised fashion.
The recurrent Grow-When-Required (GWR) network~[8] can dynamically grow or shrink in response to the input distribution.
New neurons will be created to better represent the input and connections between neurons will develop according to competitive Hebbian learning, i.e. neurons that are activated simultaneously will be connected to each other.
The network is composed of a dynamic set of neurons, $A$, with each neuron consisting of a weight vector $\textbf{w}_j$ and a number $K$ of context descriptors $\textbf{c}_{j,k}$~($\textbf{w}_j,\textbf{c}_{j,k}\in\mathbb{R}^n$).
Given the input $\textbf{x}(t)\in\mathbb{R}^n$, the index of the best-matching unit (BMU), $b$, is computed as:
\begin{equation} \label{eq:GetB}
b = \arg\min_{j\in A}(d_j),
\end{equation}
\begin{equation} \label{eq:BMU}
d_j = \alpha_0 \Vert \textbf{x}(t) - \textbf{w}_j  \Vert^2 + \sum_{k=1}^{K}\ \alpha_k \Vert \textbf{C}_k(t)-\textbf{c}_{j,k}\Vert^2,
\end{equation}
\begin{equation}\label{eq:MergeStep}
\textbf{C}_{k}(t) = \beta \cdot \textbf{w}_b^{t-1}+(1-\beta) \cdot \textbf{c}_{b,k}^{t-1},
\end{equation}
where $\Vert \cdot \Vert^2$ denotes the Euclidean distance, $\alpha_i$ and $\beta$ are constant values that modulate the influence of the temporal context, $\textbf{w}_b^{t-1}$ is the weight vector of the BMU at $t-1$, and $\textbf{C}_{k}\in\mathbb{R}^n$ is the global context of the network with $\textbf{C}_{k}(t_0)=0$.
If $K=0$, then Eq.{\ref{eq:BMU}} resembles the learning dynamics of the standard GWR without temporal context~[12].
For a given input $\textbf{x}(t)$, the activity of the network, $a(t)$, is defined in relation to the distance between the input and its BMU (Eq.~\ref{eq:GetB}) as follows:
\begin{equation} \label{eq:Activity}
a(t)=\exp(-d_b),
\end{equation}
thus yielding the highest activation value of $1$ when the network can perfectly match the input sequence~($d_b=0$).
Each neuron is equipped with a habituation counter $h_j \in [0,1]$ expressing how frequently it has fired based on a simplified model of how the efficacy of a habituating synapse reduces over time.
Newly created neurons start with $h_j=1$, with the habituation counter of the BMU, $b$, and its neighboring neurons, $n$, iteratively decreased towards 0.
The habituation rule~[12] for habituating a neuron $i$ is given by:
\begin{equation}\label{eq:FiringCounter}
\Delta h_i=\tau_i \cdot \kappa \cdot (1-h_i)-\tau_i,
\end{equation}
with $i\in\{b,n\}$ and where $\tau_i$ and $\kappa$ are constants that control the monotonically decreasing behavior.
Typically, $h_b$ is habituated faster than $h_n$ by setting $\tau_b>\tau_n$.
The network is initialized with two neurons, which is the minimum required for the discriminative ability.

At each learning iteration, a new neuron is created whenever the activity of the network, $a(t)$, in response to the input $\textbf{x}(t)$ is smaller than a given insertion threshold $a_T$.
Furthermore, $h_b$ must be smaller than a habituation threshold $h_T$ in order for the insertion condition to hold, thereby fostering the training of existing neurons before new ones are added.
The new neuron is created halfway between the BMU and the input.
The training of the existing neurons is carried out by adapting the BMU $b$ and the neurons $n$ to which the BMU is connected:
\begin{equation}\label{eq:UpdateRateW}
\Delta \textbf{w}_i = \epsilon_i \cdot h_i \cdot (\textbf{x}(t) - \textbf{w}_i),
\end{equation}
\begin{equation}\label{eq:UpdateRateC}
\Delta \textbf{c}_{i, k} = \epsilon_i \cdot h_i \cdot (\textbf{C}_k(t) - \textbf{c}_{i, k}),
\end{equation}
with $i\in\{b,n\}$ and where $\epsilon_i$ is a constant learning rate ($\epsilon_n<\epsilon_b$).
Furthermore, the habituation counters of the BMU and the neighboring neurons are updated according to Eq.~\ref{eq:FiringCounter}.
Connections between neurons are updated on the basis of neural co-activation, i.e. when two neurons fire together, a connection between them is created if it does not yet exist.

Our previously reported experiments~[9] show that growing self-organizing networks outperform a variety of proposed supervised lifelong learning approaches~[5,10,13,14] evaluated on the CORe50 dataset~[10] (see Table~1 and 2 in Supplementary Material for a quantitative comparison with other approaches).
The performance improved partly due to the recurrent learning dynamics of the networks that better capture the spatiotemporal structure of the input sequences and the replay of statistically significant neural activity patterns to mitigate the catastrophic forgetting during incremental learning.
The latter mechanism relies on the periodic re-activation of previously learned neural patterns.
This allows the consolidation of internal representations in the absence of external sensory input.
However, an empirical assessment of the contribution of neurogenesis to the overall performance has not been conducted.
(For details on the generation of neural activation trajectories and input sequence classification, see Sec.~5 in Supplementary Material.)

\section{Experiments}

We performed a series of experiments with two different models, a growing self-organizing network and a static self-organizing network (see Fig. 2.A in Supplementary Material).
There are no separate training and test phases, i.e. the model learns and triggers behavior at the same time. For our catastrophic forgetting experiments, sequential input becomes incrementally available over time and direct access to previously encountered data samples is restricted as each data sample is seen only once.
The maximum number of neurons for both models is equal, i.e. the growing network starts with two neurons and incrementally grows (as described in Sec.~2) with $N_{\textit{max}}$ being the upper-bound number of neurons, whereas the static network will be trained with a fixed number $N_{\textit{max}}$ of randomly initialized neurons.

We evaluated the performance of the models on the CORe50 dataset for continuous object recognition~[10].
The CORe50 comprises 50 objects (instances) within 10 categories with image sequences captured in 11 different sessions containing multiple views of the same objects, varying background, object pose, and degree of occlusion~(Fig.~2.B).
We used $128 \times 128$ RGB images at a frame rate of 5hz.
For a direct comparison with previously reported experiments~[9,~10], the feature extraction module consists of a VGG model pre-trained on the ILSVRC-2012 dataset to which we applied a convolutional operation with 256 1x1 kernels on the output of the fully-connected hidden layer, reducing its dimensionality from 2048 to 256.
Therefore, the self-organizing network receives a 256-dimensional feature vector per sequence frame.
We conducted our experiments on the task of instance-level object classification, using the samples from sessions $\#3$, $\#7$, $\#10$ for testing and the samples from the remaining 8 sessions for training.

For comparison, we first trained static and dynamic networks with standard iid-sampled batches, increasing $N_{\textit{max}}$ from 500 to 2500 in intervals of 100.
In general, a greater number of neurons corresponded to a better overall accuracy.
For details see Table 3 and Figure 3.
These experiments demonstrate that the static networks took longer to achieve competitive performance compared to dynamic networks (33 epochs versus 16), because the latter are able to add new neurons initialized between the input and BMUs, i.e. quickly interpolating the input data instead of needing to be iteratively adapted.
Fast insertion does not lead to catastrophic interference with the learned representation since at the point of insertion, the rest of the neural representation does not depend on the weights of the novel neuron (see [9] for an extensive discussion).

\begin{figure*}[t]
\centering
\includegraphics[width=0.8\textwidth]{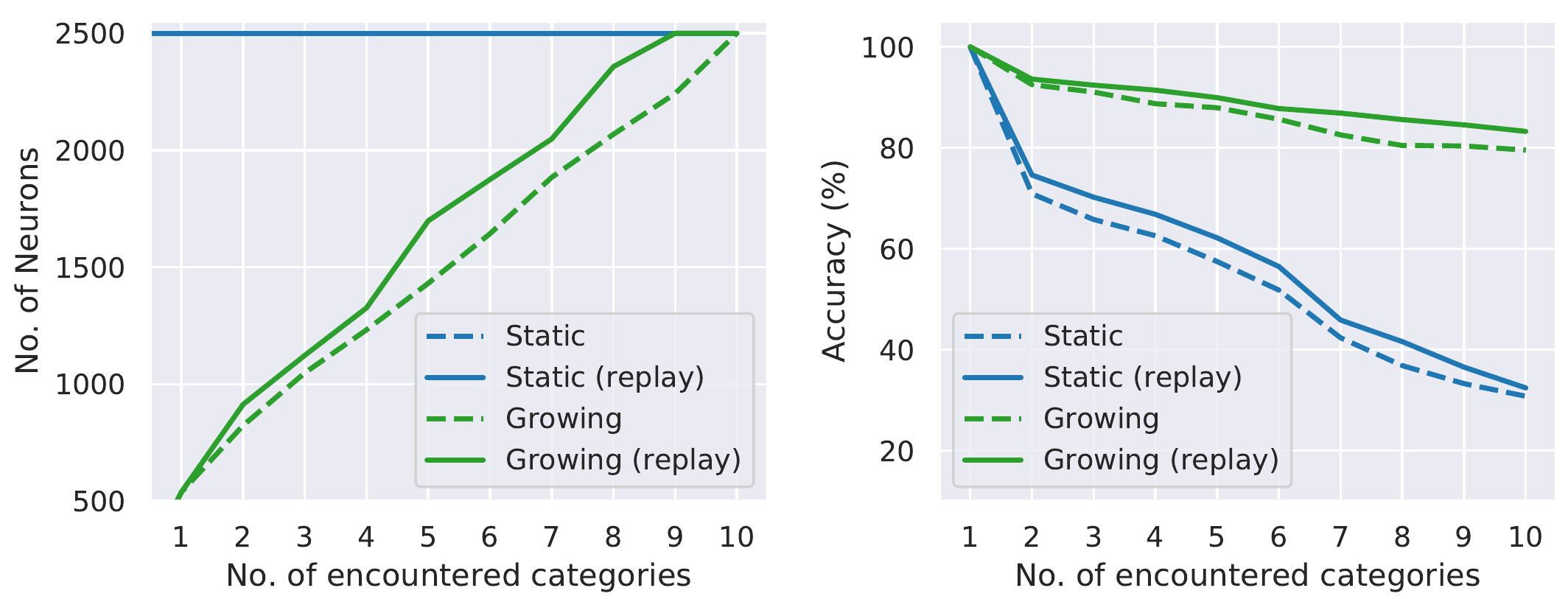}
\caption{Incremental learning of object instances with a static and a growing model. Accuracy averaged across 10 learning trials with randomly shuffled training sessions.}% (\emph{$G^C$}).}
\label{fig:results}
\end{figure*}

In the incremental learning experiment, we trained the models with mini-batches containing 5 instances per category, with one training iteration per mini-batch (i.e. previously encountered mini-batches are not shown again to the model).
We trained the models with and without memory replay and set $N_{\textit{max}}=2500$ for a comparison with the batch learning strategy.
Fig.~1 shows the average number of neurons and overall accuracy across 10 trials of randomly shuffled acquisition sessions for the incremental learning task.
For both models, the overall accuracy decreases as the number of encountered categories increases.
However, the static model shows the worst performance since one training epoch per mini-batch is not sufficient to consolidate the representations.
While in the batch learning strategy the static model required a larger number of epochs to achieve a similar performance to the dynamic model, in the incremental case the former leads to significant disruptive interference as the number of encountered categories increases.
After having encountered 10 categories, the static model shows an overall accuracy of $32.41\%$ with respect to the $83.24\%$ achieved by the dynamic model (i.e. a difference of approx. $50\%$).
The use of memory replay did not significantly improve the performance of the static network.
This is because memory replay relies on the retrieval of neural activation trajectories which, in the case of the static network, do not represent the input well.

\section{Open Issues and Future Work}

We showed that self-organizing networks with additive neurogenesis show a better performance than a static network with the same number of neurons, thereby providing insights into the design of neural architectures in incremental learning scenarios when the total number of neurons is fixed.
However, there are additional research directions to complement this study.
First, we have focused on the process of neurogenesis whereas we have not investigated the dynamic development of synaptic patterns.
Neurophysiological studies evidence that network connectivity is a crucial parameter because the ability of a network to retain memories scales in proportion to the number of synapses per neuron [11].
Consequently, one could quantitatively assess the contribution of different degrees of network connectivity and representational sparsity.
Second, we have considered additive neurogenesis without investigating the aging and removal of rarely activated neurons and connections.
This aspect is crucial for allowing to forget (not catastrophically) in networks of limited size.
Finally, additional theories suggest that one function of neurogenesis in the mammalian hippocampus is the encoding of time for the formation of temporal associations in memory~[17].
While it has been shown that unsupervised multisensory binding is possible through the use of unisensory self-organizing networks, e.g., from co-occurring audio-visual input~[18], the use of neurogenesis for temporal memory binding remains to be investigated.

\subsubsection*{Acknowledgments}

This research was partially supported by the German Research Foundation (DFG) under project Transregio Crossmodal Learning (TRR 169).

\section*{References}

\small

[1] Thrun, S. and Mitchell, T.M. (1995). Lifelong robot learning. Robot. Auton. Syst., 15:25--46.

[2] Parisi, G.I., Kemker, R., Part, J.L., Kanan, C., and Wermter, S. (2018). Continual Lifelong Learning with Neural Networks: A Review. arXiv preprint arXiv:1802.07569.

[3] Kumaran, D., Hassabis, D., and McClelland, J. L. (2016). What learning systems do intelligent agents need? Complementary learning systems theory updated. Trends in Cognitive Sciences, 20(7):512--534.

[4] Mermillod, M., Bugaiska, A. and Bonin, P. (2013). The Stability-Plasticity Dilemma: Investigating the Continuum from Catastrophic Forgetting to Age-Limited Learning Effects. Frontiers in Psychology, 4(504).

[5] Kirkpatrick, J., Pascanu, R., Rabinowitz, N., Veness, J., Desjardins, G., Rusu, A. A., Milan, K., Quan, J., Ramalho, T., Grabska-Barwinska, A., Hassabis, D., Clopath, C., Kumaran, D., and Hadsell, R. (2017). Overcoming catastrophic forgetting in neural networks. PNAS, 114(13).

[6] Zenke, F., Poole, B., and Ganguli, S. (2017). Continual learning through synaptic intelligence. ICML’17, pages 3987--3995.

[7] Rusu, A. A., Rabinowitz, N. C., Desjardins, G., Soyer, H., Kirkpatrick, J., Kavukcuoglu, K., Pascanu, R., and Hadsell, R. (2016). Progressive Neural Networks. arXiv:1606.04671.

[8] Parisi, G. I., Tani, J., Weber, C. and Wermter, S. (2017). Lifelong Learning of Humans Actions with Deep Neural Network Self-Organization. Neural Networks, 96:137--149.

[9] Parisi, G. I., Tani, J., Weber, C. and Wermter, S. (2018). Lifelong Learning of Spatiotemporal Representations with Dual-Memory Recurrent Self-Organization. arXiv preprint arXiv:1805.10966.

[10] Lomonaco, V. and Maltoni, D. (2017). CORe50: A New Dataset and Benchmark for Continuous Object Recognition. CoRL'17.

[11] Knoblauch, A. (2017). Impact of structural plasticity on memory formation and decline. In: A. van Ooyen, M. Butz (eds.): Rewiring the Brain: A Computational Approach to Structural Plasticity in the Adult Brain, 17:361--386, Elsevier/Academic Press, London, UK.

[12] Marsland, S., Shapiro, J., and Nehmzow, U. (2002). A self-organising network that grows when required. Neural Networks, 15(8–9):1041--1058.

[13] Li, Z. and Hoiem, D. (2016). Learning without forgetting. ECCV'16, pages 614--629.

[14] Rebuffi, S.-A., Kolesnikov, A., Sperl, G., and Lampert, C. H. (2017). iCaRL: Incremental Classifier and Representation Learning. CVPR’14, pages 2001--2010.

[15] Kemker, R. and Kanan, C. (2018). FearNet: Brain-Inspired Model for Incremental Learning. ICLR’18.

[16] Rusu, A. A., Rabinowitz, N. C., Desjardins, G., Soyer, H., Kirkpatrick, J., Kavukcuoglu, K., Pascanu, R., and Hadsell, R. (2016). Progressive Neural Networks. arXiv:1606.04671.

[17] Aimone, J. B., Wiles, J., Gage, F. H. (2009). Computational influence of adult neurogenesis on memory encoding. Neuron, 61:187--202.

[18] Parisi, G. I., Tani, J., Weber, C., Wermter, S. (2016). Emergence of Multimodal Action Representations from Neural Network Self-Organization. Cognitive Systems Research, 43:208--221.

%[1] Alexander, J.A.\ \& Mozer, M.C.\ (1995) Template-based algorithms
%for connectionist rule extraction. In G.\ Tesauro, D.S.\ Touretzky and
%T.K.\ Leen (eds.), {\it Advances in Neural Information Processing
%  Systems 7}, pp.\ 609--616. Cambridge, MA: MIT Press.
%
%[2] Bower, J.M.\ \& Beeman, D.\ (1995) {\it The Book of GENESIS:
%  Exploring Realistic Neural Models with the GEneral NEural SImulation
%  System.}  New York: TELOS/Springer--Verlag.
%
%[3] Hasselmo, M.E., Schnell, E.\ \& Barkai, E.\ (1995) Dynamics of
%learning and recall at excitatory recurrent synapses and cholinergic
%modulation in rat hippocampal region CA3. {\it Journal of
%  Neuroscience} {\bf 15}(7):5249-5262.

\section*{Supplementary Material}

\section{Recurrent GWR and Memory Replay}

For the purpose of memory replay in the absence of external sensory input, we generate pseudo-patterns in terms of temporally-ordered trajectories of neural activity~[9].
We use asymmetric temporal links learned by the network to recursively reactivate sequence-selective neural activity trajectories (RNATs).
We implemented temporal connections as synaptic links that are incremented between consecutively activated neurons~[18].
When the two neurons $i$ and $j$ are activated at time $t-1$ and $t$ respectively, their temporal synaptic link $P_{(i,j)}$ is increased by $\Delta P_{(i,j)} = 1$.
For each neuron $i\in A$, we can retrieve the next neuron $v$ of a prototype trajectory by selecting $v=\arg\max_{j \in A \setminus {i}} P_{(i,j)}$.
After each learning episode, i.e. a learning epoch over a mini-batch of sensory observations, we generate a RNAT, $S_j$, of length $\lambda=K+1$ for each neuron $j$ as follows:
\begin{equation}\label{eq:RNATs}
S_j=< \textbf{w}_{s(0)},\textbf{w}_{s(1)},...,\textbf{w}_{s(\lambda)} >,
\end{equation}
\begin{equation}\label{eq:RNATs1}
s(i) = \arg\max_{n \in A \setminus {j}} P_{(n,s(i-1))}, i \in [1,\lambda],
\end{equation}
where $P_{(i,j)}$ is the matrix of temporal synapses and $s(0)=j$.
The class labels of the pseudo-patterns in $S_j$ can be retrieved according to Eq.~\ref{eq:WinnerLabel}.

For the task of classification, an associative matrix $H(j,l)$ stores the frequency-based distribution of sample labels during the learning phase so that each neuron $j$ stores the number of times that an input with label $l$ had $j$ as its BMU.
The predicted label $\xi_j$ for a neuron $j$ can be computed as:
\begin{equation}\label{eq:WinnerLabel}
\xi_j = \arg\max_{l \in L} H(j,l),
\end{equation}
where $l$ is an arbitrary label.
Therefore, the unsupervised network can be used for classification without requiring the number of label classes to be predefined.

\begin{figure*}[h!]
\centering
\includegraphics[width=1\textwidth]{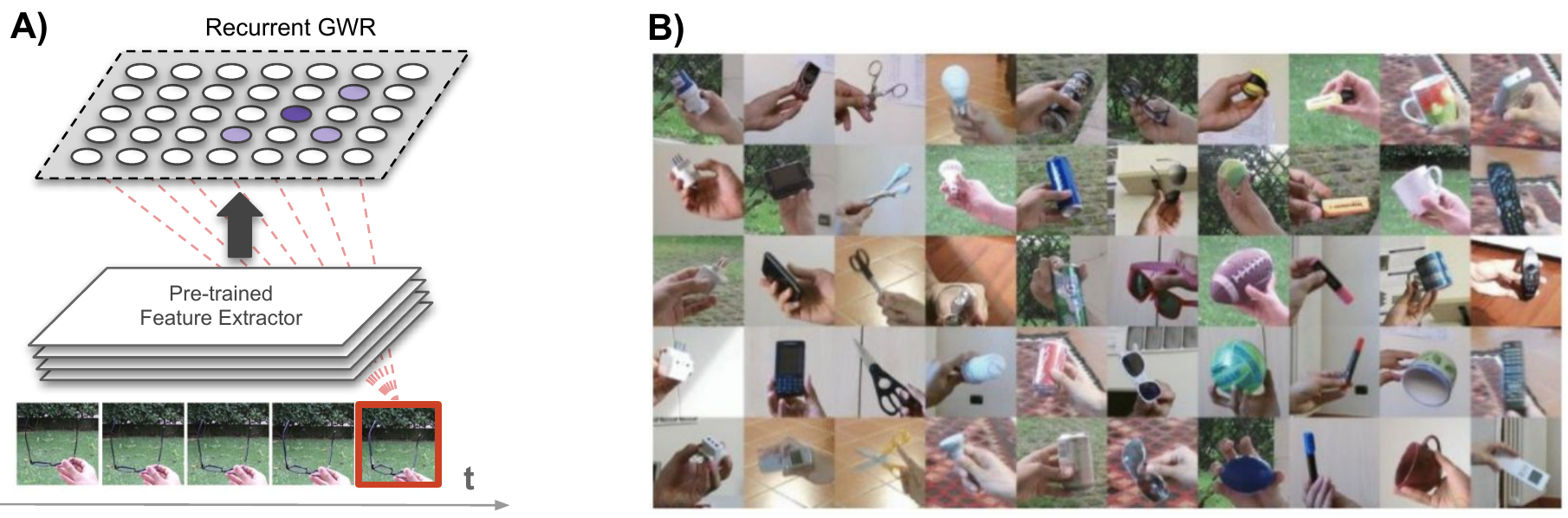}
\caption{A) Neural architecture with a deep feature extractor and a recurrent GWR (static or dynamic). B) Example frames from the CORe50 dataset showing the 10 categories (columns) comprising 5 object instances each. Adapted from [10].}% (\emph{$G^C$}).}
\label{fig:core50}
\end{figure*}

\begin{figure*}[t]
\centering
\includegraphics[width=0.75\textwidth]{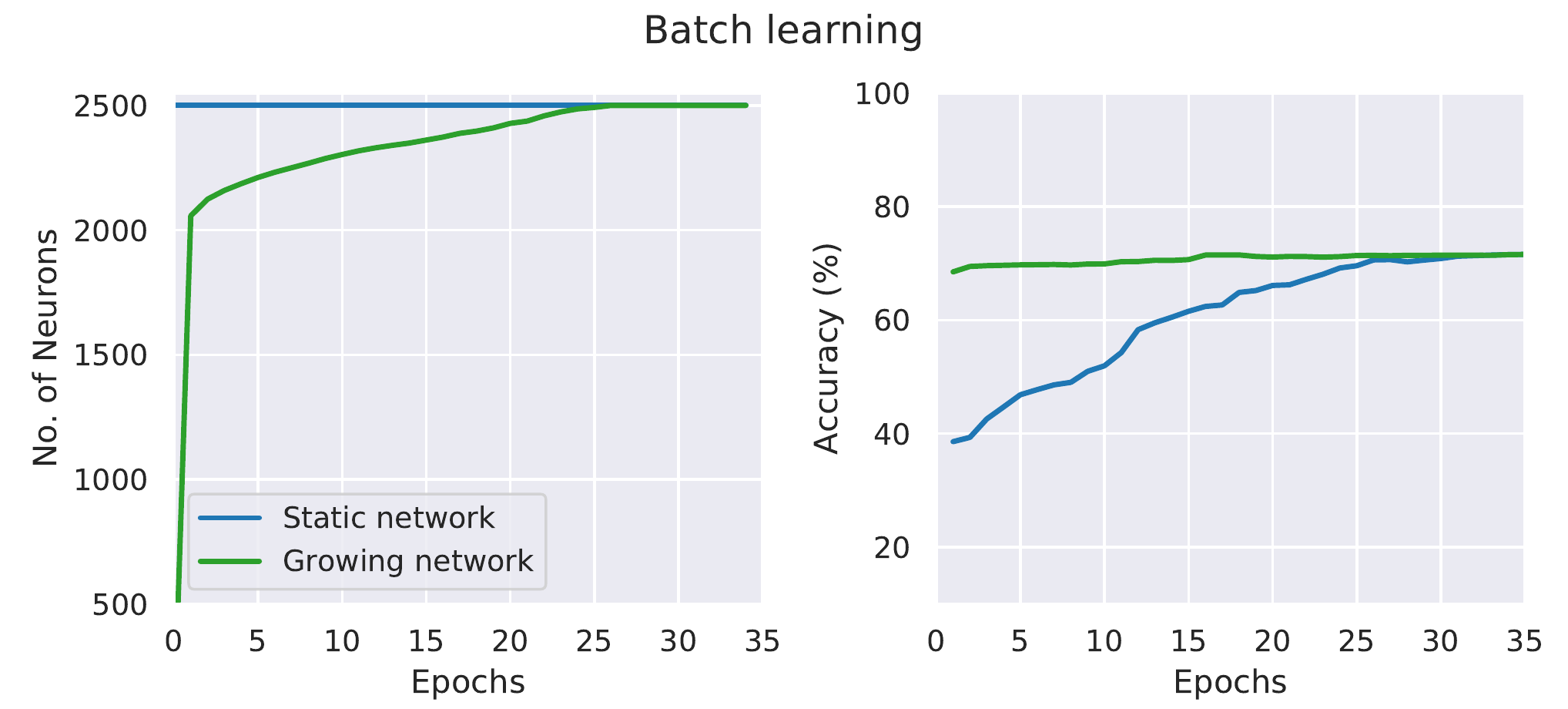}
\caption{Batch learning of instance-level classification with a static and a growing model ($N_{\textit{max}}=2500$) over 35 training epochs.}% (\emph{$G^C$}).}
\label{fig:results}
\end{figure*}

\begin{table}[h!]
\caption{Comparison of batch learning performance for instance-level classification. We show the accuracy for the pre-trained VGG with fine-tuning (VGG+FT) and the Growing Self-Organizing Memory for two different configurations: growing networks with temporal context (TC) and without TC. Best results in bold. Adapted from [9].}
\label{Tab:comparison}
\begin{center}
\begin{tabular}{lccccc}\toprule
& \textbf{Accuracy (\%)} \\
\textbf{Approach} & (Instance level: 50 classes) \\\midrule
VGG + FT [10] & 69.08 \\
Growing Self-Organizing Memory (No TC) [9] & 70.42 \\
Growing Self-Organizing Memory [9] & \textbf{79.43} \\\midrule
\end{tabular}
\end{center}
\end{table}

\begin{table}[h!]
\caption{Accuracy results with the CORe50 on 3 incremental learning scenarios. Approaches denoted with * indicate that the results for that method are based on its re-implementation by [10].}
\label{Tab:nics}
\begin{center}
\begin{tabular}{llcccc}\toprule

\textbf{Method} & \textbf{Avg. Acc. (\%)} & \textbf{Std. Dev. (\%)} \\\midrule
\textit{New Instances (NI)}  \\\midrule
Growing Self-Organizing Memory (with replay) [9] & \textbf{87.94} & 1.72\\
Growing Self-Organizing Memory [9] & 74.87 & 2.54 \\
Cumulative [19] & 65.15 & 0.66\\
LwF* [13] & 59.42 & 2.71\\
EWC* [5] & 57.40 & 3.80\\
Na\"ive [10] & 54.69 & 6.18\\\midrule
\textit{New Classes (NC) } &  \\\midrule
Growing Self-Organizing Memory (with replay) [9] & \textbf{86.14} & 2.03\\
Growing Self-Organizing Memory [9] & 73.02 & 2.91 \\
Cumulative & 64.65 & 1.04\\
iCaRL* [14] & 43.62 & 0.66\\
CWR [10] & 42.32 & 1.09\\
LwF* & 27.60 & 1.70\\
EWC* & 26.22 & 1.18\\
Na\"ive & 10.75 & 0.84\\\midrule
\textit{New Instances and Classes (NIC) } &  \\\midrule
Growing Self-Organizing Memory (with replay) [9] & \textbf{87.06} & 2.13\\
Growing Self-Organizing Memory [9] & 72.57 & 2.96 \\
Cumulative & 64.13 & 0.88\\
CWR & 29.56 & 0\\
LwF* & 28.94 & 4.30\\
EWC* & 28.31 & 4.30\\
Na\"ive & 19.39 & 2.90\\\midrule
%Training epochs & $50$ \\\midrule
\end{tabular}
\end{center}
\end{table}

\begin{table}[t!]
\caption{Training hyperparameters for the static and dynamic recurrent GWR networks.}
\label{Tab:parameters}
\begin{center}
\begin{tabular}{llcccc}\toprule
\textbf{Hyperparameters} & \textbf{Value} \\\midrule
Insertion thresholds & $a_T=0.3$ \\
Habituation counters & $h_T=0.1$, $\tau_{b}=0.3$, $\tau_{n}=0.1$, $\kappa=1.05$ \\
Temporal depth & $K=2$\\
Temporal context & $\alpha=[0.67,0.24,0.09]$, $\beta=0.7$ \\
Learning rates & $\epsilon_b=0.5$, $\epsilon_n=0.005$ \\\midrule
%Training epochs & $50$ \\\midrule
\end{tabular}
\end{center}
\end{table}

\end{document}